\documentclass[10pt,twocolumn,letterpaper]{article}

\usepackage{cvpr}
\usepackage{times}
\usepackage{epsfig}
\usepackage{graphicx}
\usepackage{amsmath}
\usepackage{amssymb}

\makeatletter
\@namedef{ver@everyshi.sty}{}
\makeatother
\usepackage{tikz}
\usetikzlibrary{shapes}
\usetikzlibrary{calc,positioning}
\usetikzlibrary{plotmarks}

\usepackage[pagebackref=true,breaklinks=true,letterpaper=true,colorlinks,bookmarks=false]{hyperref}

\cvprfinalcopy 

\def\cvprPaperID{4704} 

\ifcvprfinal\fi
\begin{document}

\title{Contrast Optimization And Local Adaptation (COALA) for HDR Compression}

\author{Shay Maymon\\
Intel\\
94 Em Hamoshavot rd, Petach Tikva, Israel\\
{\tt\small shay.maymon@intel.com}
\and
Hila Barel\\
Intel\\
94 Em Hamoshavot rd, Petach Tikva, Israel\\
{\tt\small hila.barel@intel.com}
}

\maketitle

\begin{abstract}
This paper develops a novel approach for high dynamic-range compression. It relies on the widely accepted assumption that the human visual system is not very sensitive to absolute luminance reaching the retina, but rather responds to relative luminance ratios. Dynamic-range compression is then formulated as a regularized optimization in which the image dynamic range is reduced while the local contrast of the original scene is preserved. Our method is shown to be capable of drastic dynamic-range compression, while preserving fine details and avoiding common artifacts such as halos, gradient reversals, or loss of local contrast.
\end{abstract}

\section{Introduction}
\label{sec:intro}

The dynamic range of a natural image is defined as the ratio of the lightest to darkest point in the image. 
Due to the limitation inherent in most digital image sensors, and to a lesser degree in film emulsions, it is not generally possible to capture the full spectral content and dynamic range of a scene in a single exposure. To cover the full dynamic range in such a scene, one can take a series of photographs of the same subject matter with different exposures and combine them to produce a high-dynamic range (HDR) image. 
Displaying HDR images is another problem. Various common display devices, such as CRTs, LCDs, and print media are limited to a small dynamic range of roughly one to two log units, whereas the dynamic range commonly found in real-world scenes fall well outside this range. The challenge posed is how to display HDR images on low-dynamic range display devices while preserving as much of their visual content. 
The conversion from real-world to display luminance is known as tone-mapping. 
Several approaches exist to transform HDR images into low-dynamic range images where the general appearance of the original is matched. 
These approaches can be classified into two categories: tone-reproduction curves (TRCs) and tone-reproduction operators (TROs). The former are transformations that adjust the intensity of each pixel using a function that is independent of the local spatial context, and the latter are context sensitive approaches which attempt to preserve local contrast by using the spatial structure of the image. Methods for tone-mapping started to appear in the early 1990s, and have since evolved rapidly with a vast number of tone-mapping operators introduced in the literature \cite{TR:93,Schlick:94,War:94a,DebevecMalik:97,Larson:97,Pattanaik:98,Pattanaik:02,Durand:02,Fattal:02,DMAC:03,Eilersten:15}. A nice overview of published work in the area of tone-mapping that focuses on tracking the evolution of tone-mapping is given in \cite{Eilersten:17}.

In this work we propose a novel approach for high dynamic-range compression. We formulate the compression problem as a least-squares (LS) optimization and propose an iterative technique for efficiently solving it. Our technique produces very compelling natural looking results and it can address some of the most challenging photographic scenarios. Our results are shown to outperform other results obtained with competing approaches. Our proposed iterative procedure also makes our approach competitive in terms of running time. 

The rest of the paper is organized as follows: Section \ref{sec:probForm} formulates the problem of dynamic-range compression as a regularized LS optimization; halos artifacts and approaches to mitigating them are discussed in Section \ref{sec:edges}; an iterative approach for solving the optimization is proposed in Section \ref{sec:ICM}; and details on the implementation together with results of our proposed method are shown in Section \ref{sec:results}. 

\section{Problem Formulation}
\label{sec:probForm}

Motivated by principles of the human visual system and in particular its sensitivity to relative rather than absolute luminance values, we aim at producing a reduced dynamic-range image whose local contrast is matched to that of the original scene. Our measure of local contrast is next defined.

\subsection{DRC as a regularized least-squares optimization}
\label{sec:DRC_optimization}
Given an HDR image whose luminance is denoted as ${\bf Y}$ and where ${\bf B} = \log{\bf Y}$ represents a crude approximation of its perceived brightness, we aim for reducing its dynamic range. Our measure of contrast ${C}_{ij}$ between pixel $i$ and pixel $j$ is defined as the logarithm of their luminance ratios, or equivalently as the difference between their brightness levels, \ie ,
\begin{eqnarray}
\label{eq:contrast_def}
{C}_{ij} &=& \log\left(Y_i/Y_j\right) = B_i-B_j.
\end{eqnarray}

Perfectly matching the contrast of the displayed image to that of the original image is not possible if dynamic-range compression is desired. Since certain ratios are more visually significant than others, in the trade-off between contrast preservation and dynamic-range compression, higher importance should be given to matching ratios that are more visually significant.
We thus propose formulating the problem of dynamic-range compression as the following regularized optimization:
\begin{eqnarray}
\label{eq:regLS_matrix_form}
\min_{\hat{\bf b} \leq {\bf 0}} \left\{||{\bf H} \hat{\bf b}-{\bf c}|||_{\bf W}^2 + \lambda ||\hat{\bf b} - {\bf r}||^2\right\}.
\end{eqnarray}
The objective function in (\ref{eq:regLS_matrix_form}) consists of two terms: the first is responsible for contrast preservation, whereas the second is a regularization term whose role is to shape the histogram of the image and reduce its dynamic-range. The amount of regularization is controlled by the tuning parameter $\lambda$ which determines the right balance of these two terms. 

The matrix ${\bf H}$ in (\ref{eq:regLS_matrix_form}) is a sparse matrix of size $M \times N_p$ where $M$ is the number of selected pairs involved in the contrast matching term, and $N_p$ is the total number of pixels in the image. Each row of ${\bf H}$ that corresponds to a pair $(i,j)$ has all its entities zeros except the $i$th and $j$th, which take the values $1$ and $-1$, respectively. The vectors $\hat{\bf b}$ and ${\bf r}$ are column vectors representing the brightness levels of the desired image and of the reference image, both formed by columns concatenation of their matrix representations ${\bf \hat{B}}$ and ${\bf R}$, respectively. The vector ${\bf c}$ is a column vector whose values represent the relative intensity ratios of all selected pairs in the HDR image. 


In designing the low dynamic-range image and to reflect our desire to preserve local contrast, more importance is given to preserving relative luminance ratios of neighboring pixels than to ratios of pixels that are farther apart. This goal can be achieved with a weighting function that is inversely proportional to the distance between pixels in the image. To further reduce the computational complexity, we match contrast ratios of adjacent pairs only. The contrast ratios of all adjacent pairs get equal importance so that the weighting matrix ${\bf W}$ reduces to an identity matrix. 

\begin{figure*}[ht]
\begin{center}
   \includegraphics[width=1\linewidth]{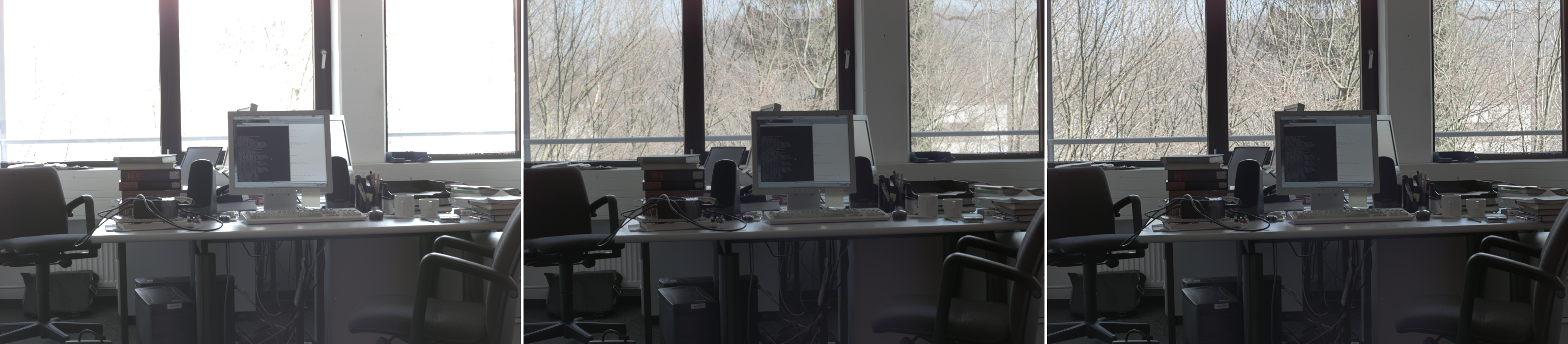}
\end{center}
   \caption{mpi office: linear scaling (left), reference image ${\bf R}$ (middle), and final compressed output ${\bf \hat{B}}$ (right). The reference image has a reduced dynamic range and the optimization adds details and enhances its local contrast, as seen in the final output (preferable viewing in a monitor).}
\label{fig:office}
\end{figure*}

The regularization term guarantees that the resulting image will be close in the LS sense to a reference image ${\bf r}$, whose dynamic range is reduced relative to that of the original HDR image. When the regularization parameter $\lambda \rightarrow \infty$, solving the optimization in (\ref{eq:regLS_matrix_form}) yields $\hat{\bf b} = {\bf r}$. As we decrease $\lambda$, the contrast of the displayed image is improved and more details are added, as illustrated in Fig. \ref{fig:office}.

The optimization yields the brightness $\hat{\bf B}$, from which the desired luminance $\hat{\bf Y}=\exp(\hat{\bf B})$ of the reduced dynamic-range image can be obtained. The optimization is constrained since ${\bf Y}$ lies in the displayable range $[0,1]$ and thus $\hat{\bf B}=\log\hat{\bf Y}$ is restricted to be non-positive. 
It is worth noting that apart from the fact that applying the logarithm function on the luminance produces an approximation of the perceived brightness, it also simplifies our model so that the unknown parameters $\{\hat{\bf B}_i\}$ are linearly related to the observations. 
The next section discusses the design of the reference image.

\subsection{Histogram adjustment through regularization}
\label{sec:regularization}

In addition to preserving the local contrast of the original HDR image, we would like to shape its histogram. Direct optimization of this form where there are constraints on the output distribution is computationally complex. This goal can be indirectly achieved by a regularization term that penalizes large deviations of the output image from a pre-specified image, whose histogram is designed as desired. For this purpose, we propose the following monotonic logarithmic TRO: 
\begin{eqnarray}
g_i(Y_i) = \frac{\log\left(Y_i+\beta {{\overline{Y}}_i}^{\gamma}\right)-\log\left(\beta {{\overline{Y}}_i}^{\gamma}\right)}{\log\left(1+\beta {{\overline{Y}}_i}^{\gamma}\right)-\log\left(\beta {{\overline{Y}}_i}^{\gamma}\right)}, \hspace{0.1in} \beta,\gamma>0.
  \end{eqnarray}
In this mapping, the parameters $\beta$ and $\gamma$ are image-dependent constants and ${\overline{Y}}_i$ is a local geometric mean of luminance values around pixel $i$.
The local-average measure ${\overline{Y}}_i$ is low for dark regions and high for bright regions of the image. This local behaviour of ${\overline{Y}}_i$ has the effect of expanding more dark regions and compressing less bright regions and it will thus preserve more details of the original scene. 
Applying this mapping on the luminance ${\bf Y}$ of the {\it HDR} image will produce a reduced-dynamic range image, whose logarithm will produce the brightness ${\bf R}$ of the reference image.


\section{Special treatment around edges}
\label{sec:edges}

\begin{figure}[h]
\begin{tikzpicture}
\node [anchor=west] (halo1) at (1.0,1.0) {};
\begin{scope}
    \node[anchor=south west,inner sep=0] (image) at (0,0) {\includegraphics[width=0.5\textwidth]{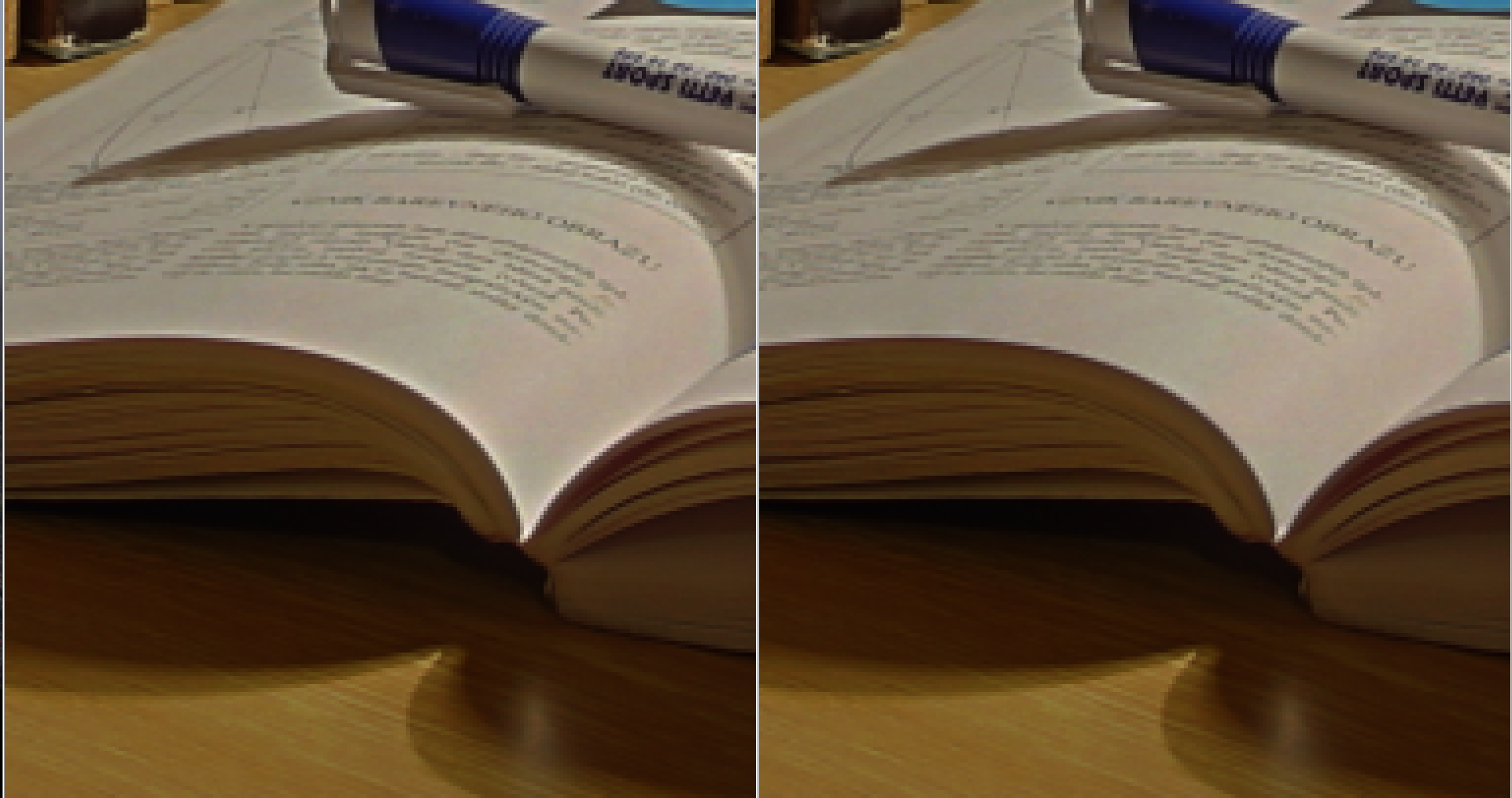}};
\end{scope}
\end{tikzpicture}
\caption{Halos reduction with local upper bound. Left: zero upper bound. Right: upper bound is edge aware. } 
\label{fig:halos}
\end{figure}

Matching the local contrast of the displayed image to that of the HDR image may reveal some difficulties around strong edges. Specifically, high local contrast around strong edges together with the unforgiving behaviour of $l_2$-norm to large errors results in over-exaggerated contrast, referred to as halos.
To reduce these halos, we propose gradually increasing the regularization parameter $\lambda$ in the vicinity of strong edges. The importance of preserving local contrast of the HDR image in these regions will  decrease relative to the importance of matching the reference image, whose contrast is reduced. 
While this approach works well in reducing dark halos, for bright halos around dark edges, a different approach is found to work better. Specifically, rather than constraining the desired brightness to be non-positive,  in the regions of strong edges we gradually reduce its upper bound from zero until it matches the reference image. These two context-dependent solutions 
significantly reduce halos artifacts. Fig.~\ref{fig:halos} illustrates halos reduction when using local upper bound. 

\section{Efficiently Solving the Optimization}
\label{sec:ICM}
The enormous size of the problem makes the constrained optimization of (\ref{eq:regLS_matrix_form})
computationally demanding, despite the sparseness of ${\bf H}$. This section introduces the Iterated Conditional Modes (ICM) method \cite{Kittler:84} as an efficient technique for solving the LS optimization. 
The ICM technique is simply an application of coordinate-wise gradient ascent. It consists of a simple local computation that can be performed
efficiently. This local computation can be repeated in a systematic way, for
instance by repeatedly raster scanning through the image, or by choosing nodes at
random, until some suitable stopping criterion is satisfied. The ICM algorithm is claimed to converge to a local optimum if no changes are made to the variables in a sequence of updates where every node is visited at least once. This need not, however, correspond to the global optimum. 
This means that ICM can be used to iteratively solve the constrained LS optimization of (\ref{eq:regLS_matrix_form}). Since each step of the procedure is just a scalar optimization, the computational cost of the iterative procedure is much cheaper than that of a direct solution. With asynchronous updating, the iterative procedure is guaranteed to converge. Since the objective function is convex it is assured to converge to the global optimum. Synchronous updating, on the other hand, has the advantage of being parallelized. 

\section{Implementation and Results}
\label{sec:results}

\begin{figure*}[p]
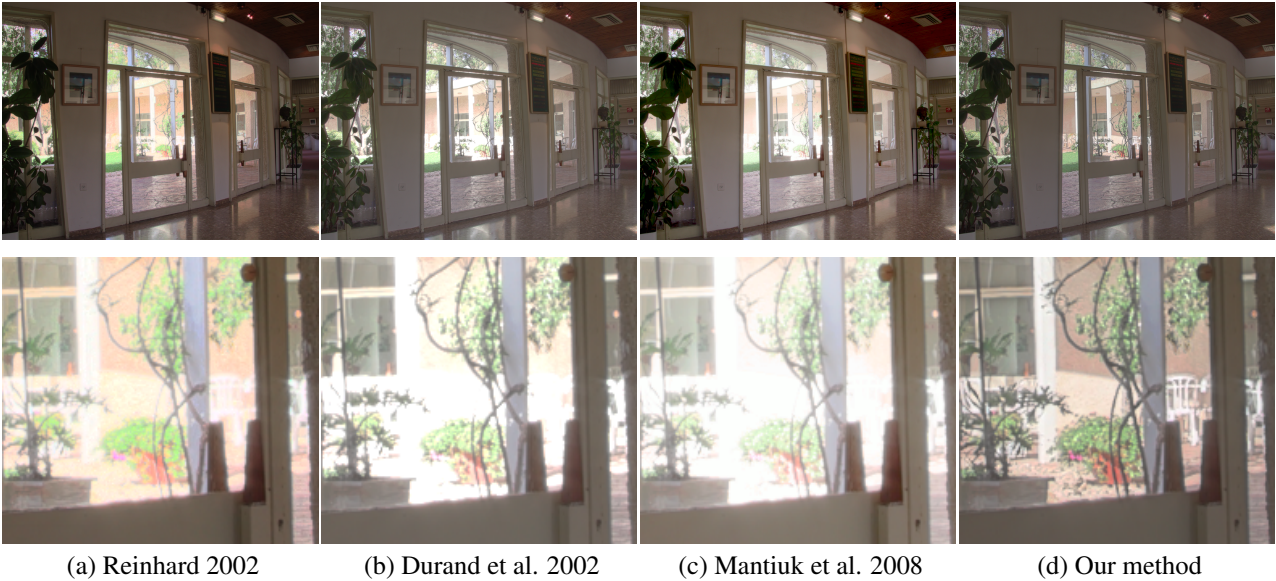

\begin{center}
\input Results_icip
\end{center}
\caption{Belgium House: full image rendition ($1$st row) and close up of the garden ($2$nd row). All operators show pleasing and natural images. (a)-(c) lose information in bright regions whereas our method renders the highlights better without sacrificing visibility at shadows.}
\label{fig:results1}
\end{figure*}

\begin{figure*}[p]
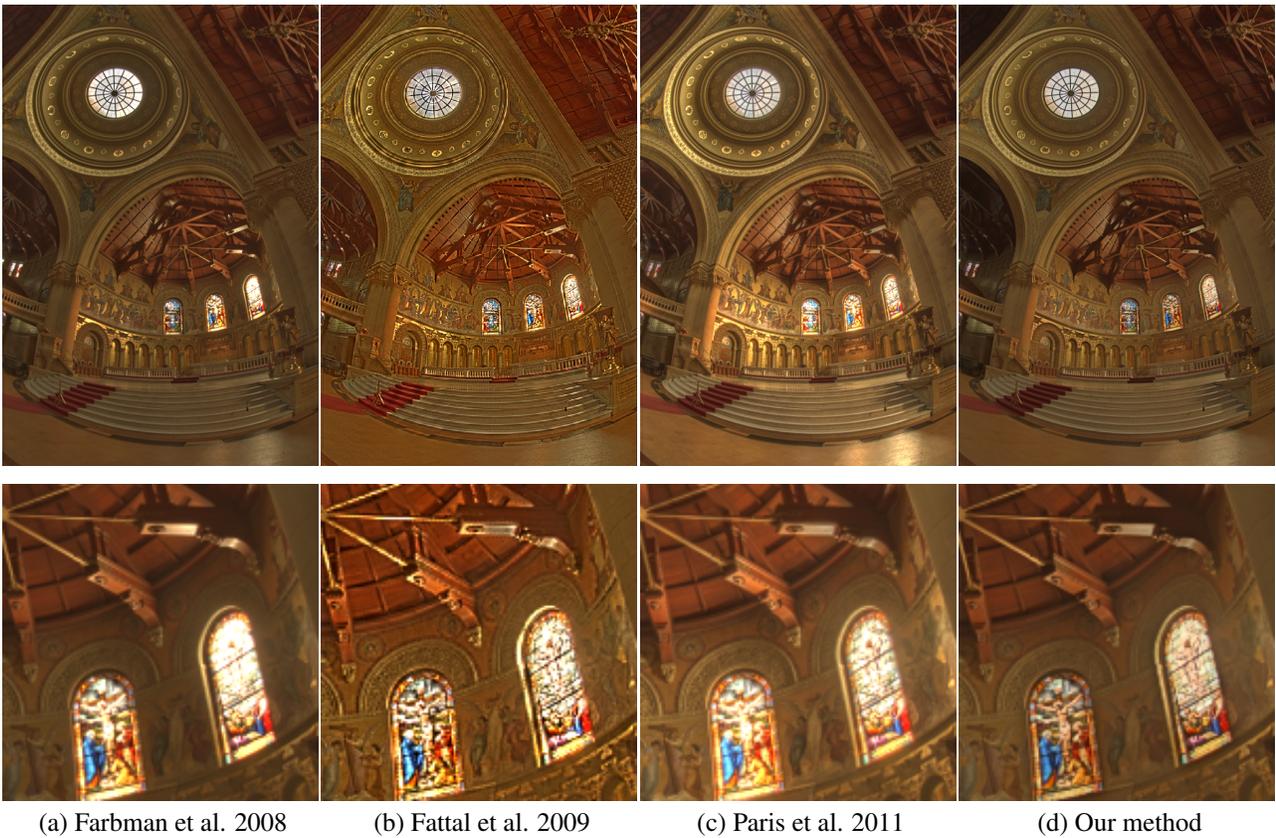

\begin{center}
\input Results_icip2
\end{center}
\caption{A comparison of our approach with multi-scale approaches for tone-mapping. In our result, as seen on the window crop, details are better recovered in highlights and the image has an overall natural look.}
\label{fig:results2}
\end{figure*} 

This section discusses the implementation details and the results achieved with our proposed algorithm.
Our method operates on the luminance of the RGB image in the linear domain and as such it keeps the colors unchanged \cite{Durand:02}. The luminance values are chosen as the maximum of the RGB triplet. This choice guarantees that none of the RGB channels will exceed the displayable range when multiplied with the gain representing the ratio between the luminance after compression and the luminance of the HDR image. 
The luminance of the compressed image is computed by iteratively solving the constrained LS optimization using ICM with synchronous updating with the reference image used as an initialization.

Our proposed algorithm was implemented in C++ using Intel's SSE intrinsic instructions. One megapixel image is processed in about 100 msec on a 3.6GHz Intel Core i7 CPU with a single thread. The computation time is linear with the size of the image and it can be speed-up further using OpenMP. Experiments show convergence of the iterative procedure to the optimal solution in less than 50 iterations. We expect to achieve faster convergence rate with better initializations. This can be utilized for video processing as well. In order to accelerate the convergence rate one can use the optimal solution of the previous frame as an initialization for the next frame. Apart from saving in computation, the iterative form of our proposed solution can provide better temporal stability and no flickering during video capture.

We have validated our method on a variety of HDR radiance images commonly used in the literature. Two are shown in figures \ref{fig:results1}, \ref{fig:results2} and the remaining images are included in the supplementary material. Our method can address some of the most challenging photographic scenarios. In all cases tested it produced natural-looking results with the appearance of the original scene preserved. We compared our algorithm with other popular methods. For this purpose, we chose a representative set of tone reproduction algorithms producing subjectively pleasing results, \ie, \cite{Fattal:02, Reinhard:02, Durand:02, Mantiuk:08, Farbman:08, Fattal:09, Paris:11}. It is worth noting that the goal of some of these operators is details enhancement as well as range compression, whereas our goal is to maintain the subjective experience with the real scene. We therefore selected only those results representing neutral image rendition.  
Below are the resulting images where their full resolution can be found in the supplementary material. In producing our results we used default tuning parameters that were found to yield satisfactory results. Processing the results was done using the pfstools package \cite{mantiuk:2007:hvei} and a standard gamma curve was used for displaying purposes.
Fig.~\ref{fig:results1} compares our proposed method with three classical tone-mapping algorithms, \ie , Durand and Dorsey \cite{Durand:02}, Reinhard \etal \cite{Reinhard:02} and Mantiuk \etal \cite{Mantiuk:08} for the challenging Belgium house HDR image. All tone-mapping algorithms show pleasing and natural images, however while \cite{Durand:02}, \cite{Reinhard:02}, and \cite{Mantiuk:08} fail to reproduce information in highlights, our proposed method succeeds to extract many details in highlights.
Fig.~\ref{fig:results2} compares our algorithm with three other multi-scale approaches, \ie , Farbman \etal \cite{Farbman:08}, Fattal \cite{Fattal:09} and Paris \etal \cite{Paris:11} for Stanford Memorial Church. The dynamic range of this image exceeds 250,000:1. Images produced by \cite{Farbman:08} and \cite{Fattal:09} appear $''$flat$''$ while \cite{Paris:11} has an unrealistic look. Details in highlights are lost in all three images, as can be seen in the small crops. While maintaining overall natural look, our method achieves better details extraction.


\section{Future work}
\label{sec:future_work}
As a future work we plan to extend this framework to apply local-contrast enhancement to non-HDR images. This can be achieved by modifying the first term in the optimization to enhance rather than preserve contrast. Enhancement should be content dependent so that contrast of flat regions will be preserved whereas contrast of detailed areas will be enhanced. We also plan to investigate a different approach for reducing  halos artifacts that adds monotonicity constraints to the optimization to avoid reversals. 
In addition, we plan to adapt the proposed approach to video sequencing and utilize the similarity between consecutive frames in the iterative procedure. 

\section{Conclusion}
\label{sec:conclusions}
This paper proposes a novel approach for high-dynamic range compression. We formulate the compression problem as a regularized least-squares optimization which consists of a contrast preserving term and a regularization term that shapes the histogram of the resulting output and reduces its dynamic range. It is shown that using ICM to solve the optimization significantly reduces the running time, and convergence to the optimal solution with synchronous updating is achieved in less than 50 iterations. Our proposed method is shown to produce natural-looking artifact-free images that maintain the subjective experience with the real scene. Comparing our results to other popular techniques commonly used to compress HDR images reveals that our approach reveals more details and look more natural.


\newpage
{\small
\bibliographystyle{ieee}
\bibliography{egbib}

\begin{thebibliography}{10}\itemsep=-1pt

\bibitem{DebevecMalik:97}
P.~E. Debevec and J.~Malik.
\newblock Recovering high dynamic range radiance maps from photographs.
\newblock {\em Proc. ACM SIGGRAPH}, page 369–378, 1997.

\bibitem{DMAC:03}
F.~Drago, K.~Myszkowski, T.~Annen, and N.~Chiba.
\newblock Adaptive algorithmic maping for displaying high contrast scenes.
\newblock {\em Computer Graphics Forum}, pages 419--426, 2003.

\bibitem{Durand:02}
F.~Durand and J.~Dorsey.
\newblock Fast bilateral filtering for the display of high-dynamic-range
  images.
\newblock {\em Proc. ACM SIGGRAPH}, 2002.

\bibitem{Reinhard:02}
P.~S. E.~Reinhard, M.~STARK and J.~FERWERDA.
\newblock Photographic tone reproduction for digital images.
\newblock {\em ACM Transactions on Graphics (TOG)}, 21(3):267–276, 2002.

\bibitem{Eilersten:17}
G.~Eilersten, R.~K. Mantiuk, and J.~Unger.
\newblock A comparative review of tone-mapping algorithms for high dynamic
  range video.
\newblock {\em Comput. Graph. Forum}, 36(2):565–592, 1917.

\bibitem{Eilersten:15}
G.~Eilersten, R.~K. Mantiuk, and J.~Unger.
\newblock Real-time noise aware tone mapping.
\newblock {\em ACM Trans. Graph}, 34(6), 2015.

\bibitem{Farbman:08}
Z.~Farbman, R.~Fattal, D.~Lischinski, and R.~Szeliski.
\newblock Edge-preserving decompositions for multi-scale tone and detail
  manipulation.
\newblock {\em Transactions on Graphics (TOG)}, 27(3), 2008.

\bibitem{Fattal:09}
R.~Fattal.
\newblock Edge-avoiding wavelets and their applications.
\newblock {\em Proc. ACM SIGGRAPH}, 2009.

\bibitem{Fattal:02}
R.~Fattal, D.~Lischinski, and M.~Werman.
\newblock Gradient domain high dynamic range compression.
\newblock {\em Proc. ACM SIGGRAPH}, 2002.

\bibitem{Kittler:84}
J.~Kittler and J.~F¨oglein.
\newblock Contextual classification of multispectral pixel data.
\newblock {\em Image and Vision Computing}, 2:13--29, 1984.

\bibitem{Larson:97}
G.~W. Larson, H.~Rushmeier, and C.~Piatko.
\newblock A visibility matching tone reproduction operator for high dynamic
  range scenes.
\newblock {\em IEEE Transactions on Visualization and Computer Graphics},
  3(4):291–306, 1997.

\bibitem{Mantiuk:08}
R.~Mantiuk, S.~Daly, and L.~Kerofsky.
\newblock Display adaptive tone mapping.
\newblock {\em Transactions on Graphics (TOG)}, 27(3), 2008.

\bibitem{mantiuk:2007:hvei}
R.~Mantiuk, G.~Krawczyk, R.~Mantiuk, and H.-P. Seidel.
\newblock High dynamic range imaging pipeline: Perception-motivated
  representation of visual content.
\newblock In B.~E. Rogowitz, T.~N. Pappas, and S.~J. Daly, editors, {\em Human
  Vision and Electronic Imaging XII}, volume 6492 of {\em Proceedings of SPIE},
  San Jose, USA, February 2007. SPIE.

\bibitem{Paris:11}
S.~Paris, S.~Hasinoff, and J.~Kautz.
\newblock Local laplacian filters: Edge-aware image processing with a laplacian
  pyramid.
\newblock {\em Transactions on Graphics (TOG)}, 30(4), 2011.

\bibitem{Pattanaik:98}
S.~Pattanaik, J.~A. Ferwarda, M.~D. Fairchild, and D.~P. Greenberg.
\newblock A multiscale model of adaptation and spatial vision for realistic
  image display.
\newblock {\em Proceedings of the 25th annual conference on computer graphics
  and interactive techniques}, pages 287--298, 1998.

\bibitem{Pattanaik:02}
S.~Pattanaik and H.~Yee.
\newblock Adaptive gain control for high dynamic range image display.
\newblock {\em In proceedings of the 18th spring conference on computer
  graphics}, pages 83--87, 2002.

\bibitem{Schlick:94}
C.~Schlick.
\newblock Quantization techniques for visualization of high dynamic range
  pictures.
\newblock {\em Springer-Verlag}, pages 7--20, 1994.

\bibitem{TR:93}
J.~Tumblin and H.~Rushmeier.
\newblock Tone reproduction for realistic images.
\newblock {\em Computer Graphics and Applications}, 13(6):42--48, 1993.

\bibitem{War:94a}
G.~Ward.
\newblock A contrast- based scalefactor for luminance display.
\newblock {\em Graphics gems}, IV:415--421, 1994.

\end{thebibliography}
}

\end{document}